\newcommand{\CLASSINPUTtoptextmargin}{1.9cm}
\newcommand{\CLASSINPUTbottomtextmargin}{2.54cm}
\documentclass[conference,a4paper]{IEEEtran}

\IEEEoverridecommandlockouts

\usepackage{multirow}
\usepackage{caption}
\usepackage{subcaption}
\usepackage{tabularx}
\usepackage{cite}
\usepackage{amsmath,amssymb,amsfonts}
\usepackage{algorithmic}
\usepackage{graphicx}
\usepackage{textcomp}
\usepackage{xcolor}
\usepackage{fancyhdr}

\begin{document}

\title{Contrast Enhancement of Medical X-Ray Image Using Morphological Operators with Optimal Structuring Element\\
{\footnotesize }
}

\author{
	\IEEEauthorblockN{Rafsanjany Kushol\IEEEauthorrefmark{1},
		Md. Nishat Raihan\IEEEauthorrefmark{2},
		A. B. M. Ashikur Rahman\IEEEauthorrefmark{3} and
		Md Sirajus Salekin\IEEEauthorrefmark{4}
	}
	\IEEEauthorblockA{
		\IEEEauthorrefmark{1}\IEEEauthorrefmark{2}\IEEEauthorrefmark{3}Department of Computer Science and Engineering, Islamic University of Technology, Dhaka, Bangladesh}
	\IEEEauthorblockA{
		\IEEEauthorrefmark{4}Department of Computer Science and Engineering, University of South Florida, Florida, United States}
	\IEEEauthorblockA{
		\IEEEauthorrefmark{1}kushol@iut-dhaka.edu,
		\IEEEauthorrefmark{2}nishatraihan@iut-dhaka.edu,
		\IEEEauthorrefmark{3}ashikiut@iut-dhaka.edu and
		\IEEEauthorrefmark{4}salekin@mail.usf.edu
	}
}

%

\maketitle


\begin{abstract}
To guide surgical and medical treatment X-ray images have been used by physicians in every modern healthcare organization and hospitals. Doctor’s evaluation process and disease identification in the area of skeletal system can be performed in a faster and efficient way with the help of X-ray imaging technique as they can depict bone structure painlessly. This paper presents an efficient contrast enhancement technique using morphological operators which will help to visualize important bone segments and soft tissues more clearly. Top-hat and Bottom-hat transform are utilized to enhance the image where gradient magnitude value is calculated for automatically selecting the structuring element (SE) size. Experimental evaluation on different x-ray imaging databases shows the effectiveness of our method which also produces comparatively better output against some existing image enhancement techniques.
\end{abstract}


\section{Introduction}
With the rapid advancement of medical images, it has become quite common to accurately identify the diseases and perform diagnosis automatically in modern healthcare system. In most of the cases, it is essential to enhance the contrast of the image captured from the machine. The enhancement methods are fundamentally a collection of techniques which are applied in order to improve the visual quality of the image. In several types of medical images, proper pathology detection is quite challenging for the human eye. Without necessary contrast enhancement, some images are not always functional. Nowadays, medical image enhancement techniques are playing more and more important roles not only for disease identification but also in critical surgery operation, pregnancy complication monitoring, severe disease screening, health management, early diagnosis etc. \par

One of the major types of medical images is X-ray image which promoted the world of medical science in an advanced stage. When X-rays pass through human body it travels through the skin easily but takes a little bit more time to cross the bone segment. Thus produce a brighter region around bone-like structure in the photographic plane. There are far too many applications of X-ray images not only in medical science but also in some real-life scenarios. The most common applications are the detection of broken bones inside human body, detection of several types of diseases, assisting doctors during surgical treatment, and the radiation therapy. However, an X-ray image of human chest is a powerful way of human identification system during mass disasters \cite{b1}. Other than medical diagnostic applications, the X-ray imaging technique is applied for the airport luggage security measures to check dangerous or harmful items as well as quality monitoring program of different packaged components. \par

Medical science recognizes different types of X-ray images. Chest X-ray (CXR) images are prescribed in case of shortness of breath, fever, chest pain etc. Furthermore, different types of lung diseases like lung cancer, pneumonia, and pulmonary edema can be confirmed by CXR report. Dental X-ray guides physicians to identify oral and teeth related diseases. Another important type is the bone X-ray which provides a high level of details about bones and supporting tissues of different parts of the body. In case of bone X-ray images, detection of broken or cracked bones and related diseases can be identified straight from the X-ray image. Joint abnormalities of different bone segments of the body like arm, knee, foot, hip, wrist, shoulder etc. can be determined from bone X-ray images. Moreover, X-ray images are also helpful to analyze pathology like kidney stones as well as orthopedic implant. \par

X-ray images are commonly grey-scale image with a high amount of noise and a low level of intensity. However, the contrast in the image and the boundary representation is comparatively poor and weak \cite{b2}. With very limited information and low-quality image, visually feature extraction from these X-ray images is quite a challenging task. The quality of these images can be enhanced by applying some contrast enhancement techniques. Thus segmentation and feature extraction from these images can be performed with more efficiency and comfort. \par

This study presents an efficient and robust contrast enhancement technique for different X-ray images using morphological operator top-hat and bottom-hat transform where the SE size is selected automatically from the gradient magnitude value. \par

\section{LITERATURE REVIEW}

Due to the huge amount of applications in the area of our lives, especially in the diagnosis of the medical diseases, the research field of medical image enhancement is a prominent one. Some research works in the area of X-ray imaging fully focused on image enhancement whereas a few papers focused on segmenting different parts or tissues after applying an image enhancement algorithm as a pre-processing step. Table \ref{tab1} shows a complete summary of all the related works. \par

There are quite some notable works on X-ray images of the chest in recent years. A two-scale Retinex with different weighted factors was implemented by Chen and Zou \cite{b3} for enhancing the contrast of CXR images. While examining an X-ray image of the chest, it is important to know from which viewing angle it is observed. Xue et al. \cite{b4} developed a methodology to detect whether a CXR image is the frontal view or the lateral view based on histogram of oriented gradients and contour based shape descriptor. A deep learning model based on Convolution Neural Network (CNN) and Recurrent Neural Network (RNN) is also designed by Shin et al. \cite{b5} to detect common diseases from CXR images as well as provide information about severity, location, and the affected organs. Rajpurkar et al. \cite{b6} developed their own algorithm CheXNet based on a 121 layered CNN which can detect radiologist-level pneumonia from X-ray images of the chest. \par

In case of dental X-ray images sometimes also referred to as DXR images, these are used to detect several types of dental diseases, cracks and the overall condition of the teeth. A comparative analysis of four prominent enhancement methods is performed by Ahmad et al. \cite{b7} for DXR images. They are adaptive histogram equalization (AHE), contrast limited adaptive histogram equalization (CLAHE), sharp contrast limited adaptive histogram equalization (SCLAHE), and median adaptive histogram equalization (MAHE). Naam et al. \cite{b8} implemented a Multiple Morphological Gradient (mMG) method on panoramic DXR images to diagnose dental decay originated from bacterial infections. On the other hand, Ngan et al. \cite{b9} designed a framework where the image of dental X-ray is segregated into some segments first and then detect the potential diseases using fuzzy aggregation operators and Affinity Propagation Clustering (APC+). \par

For X-ray images of bones, it is important to enhance the quality of the image as it consists of a lot of critical information in it. An image enhancement technique based on image fusion using a discrete wavelet transform and Wiener filter is developed by Khan et al. \cite{b10} for X-ray images. Yijiang Zhang \cite{b11} implemented Fruit Fly optimization algorithm to enhance the quality of the medical X-ray image. Huang et al. \cite{b12} suggested a noise removal and contrast enhancement method based on two-stage filtering, adaptive filtering and bilateral filtering. They improved the contrast by applying grey-level morphology and CLAHE. On the other hand, Rui and Guoyu \cite{b13} used a different type of filter called TV-Homomorphic filter which was able to balance the brightness and enhancing the detail information. \par

\begin{table} 
	\centering
	\caption{SUMMARY OF RELATED WORKS}
	\label{tab1}
	\begin{tabular}{|c|c|c|c|}
		\hline
		&  &  &  \\
		\textbf{X-Ray} & \multirow{2}{*}{\textbf{Work}} & \multirow{2}{*}{\textbf{Year}} & \multirow{2}{*}{\textbf{Method Used}} \\
		\textbf{Type} &  &  &  \\
		 &  &  &  \\
		\hline
		\multirow{8}{*}{Chest} & Chen  &  \multirow{2}{*}{2009} & Two-scale Retinex with   \\
		& and Zou \cite{b3} &  & different weighted factors \\
		\cline{2-4}
		&  & & Histogram of Oriented   \\
		& Xue et al. \cite{b4} & 2015  & Gradients and Contour-based \\
		&  &  & shape descriptor \\
		\cline{2-4}
		&  \multirow{2}{*}{Shin et al. \cite{b5}} &  \multirow{2}{*}{2016} & CNN and Recurrent  \\
		&  &  & Neural Network (RNN) \\
		\cline{2-4}
		& Rajpurkar et  &  \multirow{2}{*}{2017} & \multirow{2}{*}{121 layered CNN} \\
		& al. \cite{b6} &  &  \\
		\hline
		\multirow{6}{*}{Dental} &  \multirow{2}{*}{Ahmad et al. \cite{b7}} &  \multirow{2}{*}{2012} & AHE, CLAHE, SCLAHE  \\
		&  &  & MAHE \\
		\cline{2-4}
		&  \multirow{2}{*}{Naam et al. \cite{b8}} &  \multirow{2}{*}{2016} & Multiple Morphological  \\
		&   &  & Gradient (mMG) \\
		\cline{2-4}
		&  &  & Fuzzy Aggregation \\
		& Ngan et al. \cite{b9}  & 2016 & Operators and Affinity\\
		&  &  &  Propagation Clustering \\
		\hline
		&  \multirow{2}{*}{Khan et al.  \cite{b10}} &  \multirow{2}{*}{2016} & Discrete Wavelet transform  \\
		&  &  &   and Wiener filter \\
		\cline{2-4}
		& Yijiang  & \multirow{2}{*}{2016} & Fruit Fly optimization \\
		Bone or &  Zhang \cite{b11} &  &   algorithm \\
		\cline{2-4}
		Mixed &  \multirow{2}{*}{Huang et al. \cite{b12}} &  \multirow{2}{*}{2016} & Adaptive filtering, Bilateral  \\
		&   &  &  filtering and CLAHE \\
		\cline{2-4}
		& Rui and  &  \multirow{2}{*}{2017} &  \multirow{2}{*}{TV-Homomorphic filter} \\
		&  Guoyu \cite{b13} &  &   \\
		\hline
	\end{tabular}
\end{table}

\section{PROPOSED METHOD}

\subsection{Apply Combined Top-hat and Bottom-hat Transform}

\begin{figure}[h]
	\centering
	\begin{subfigure}[b]{0.5\linewidth}
		\centering\includegraphics[width=4.2cm,height=4.6cm]{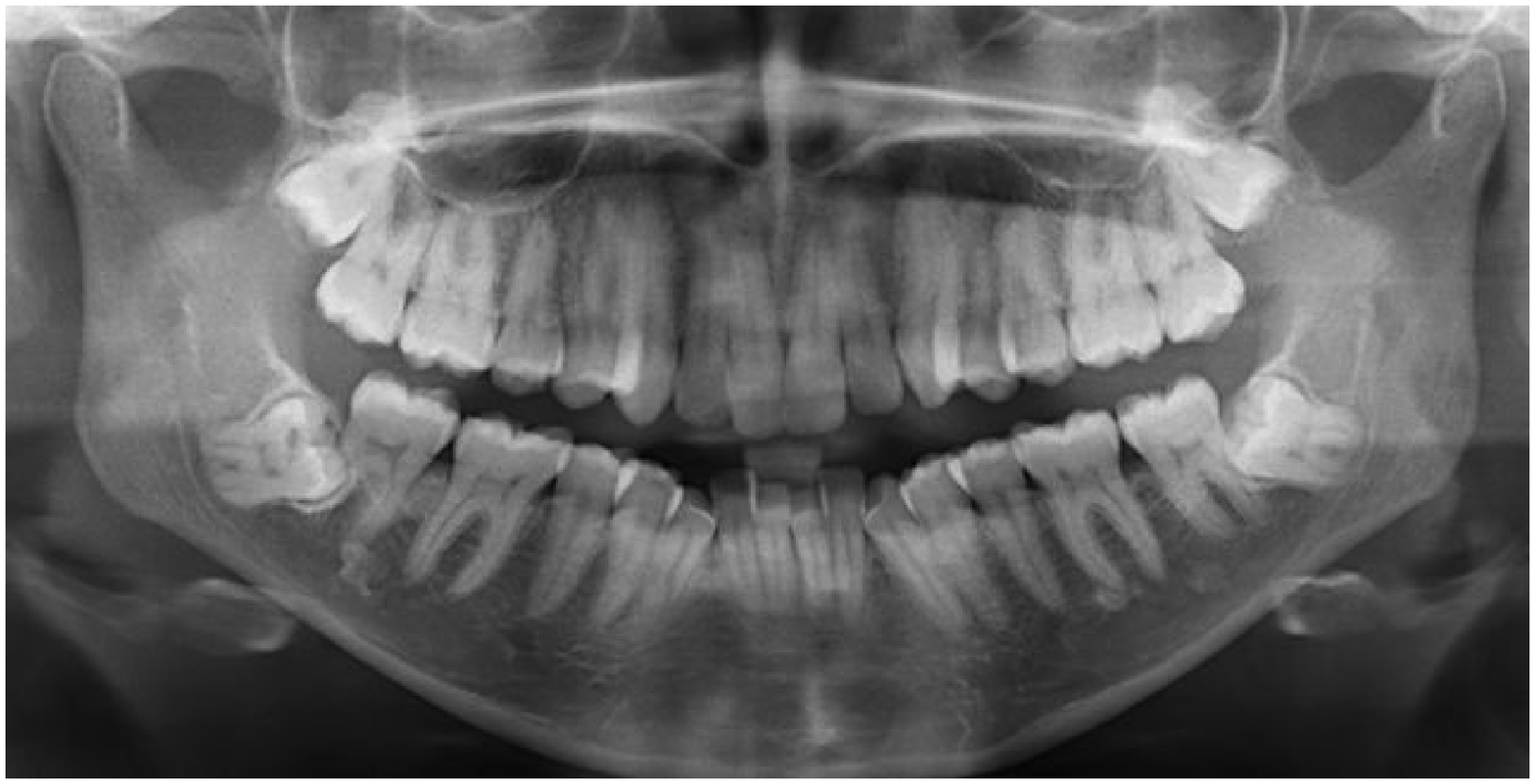}
		\caption{}
	\end{subfigure}%
	\begin{subfigure}[b]{0.5\linewidth}
		\centering\includegraphics[width=4.2cm,height=4.6cm]{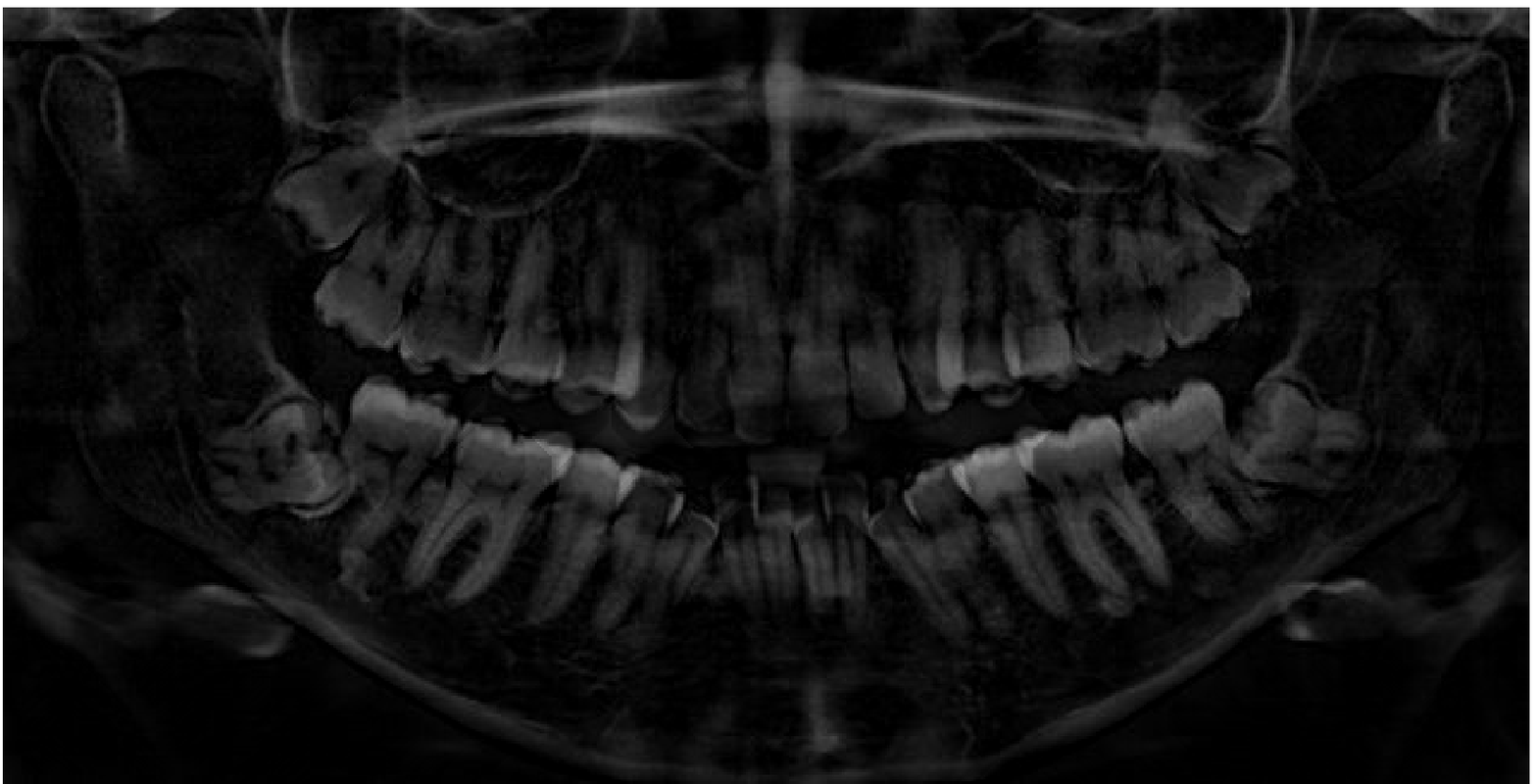}
		\caption{}
	\end{subfigure} \\
	\centering
	\begin{subfigure}[b]{0.5\linewidth}
		\centering\includegraphics[width=4.2cm,height=4.6cm]{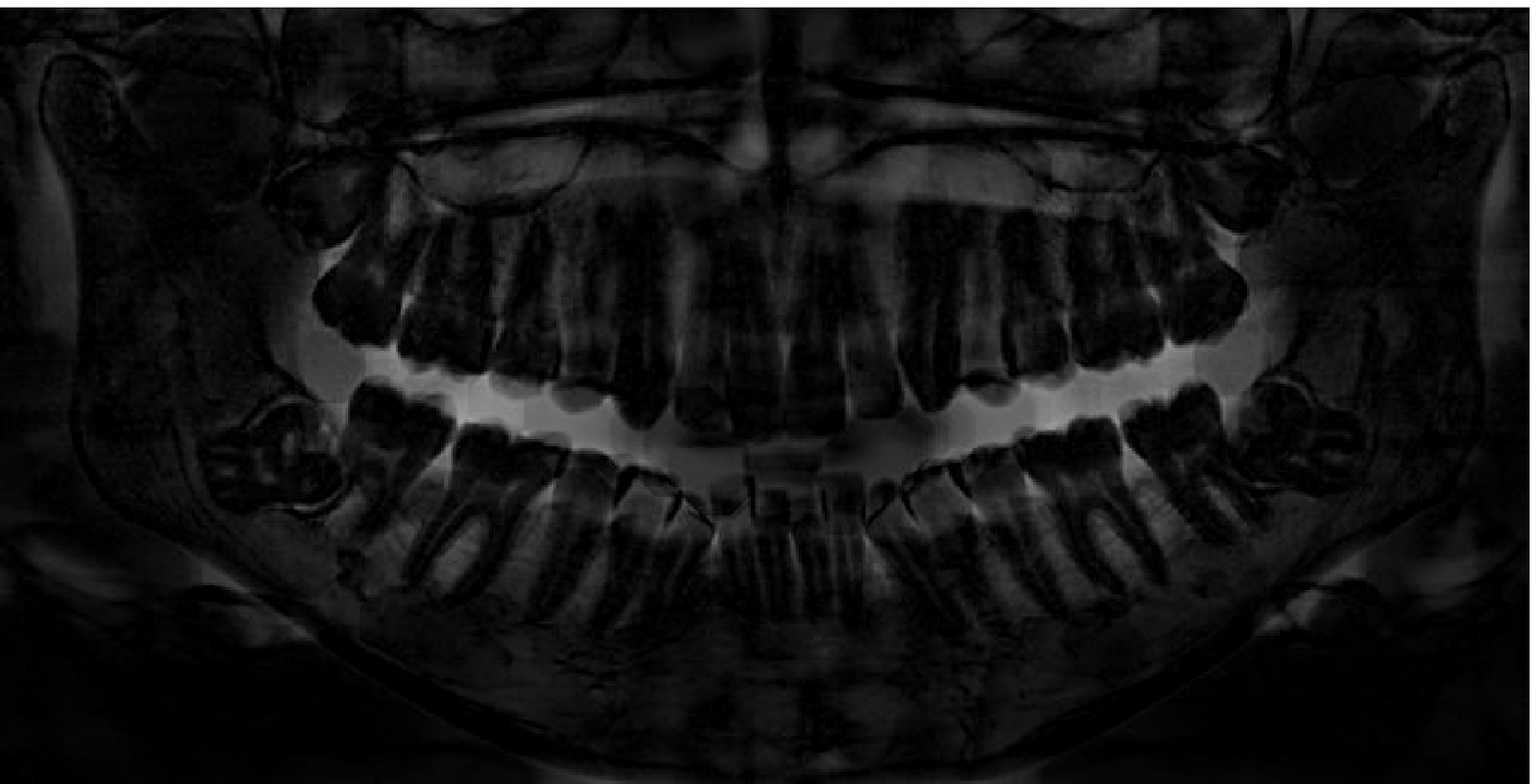}
		\caption{\label{fig:fig1}}
	\end{subfigure}%
	\begin{subfigure}[b]{0.5\linewidth}
		\centering\includegraphics[width=4.2cm,height=4.6cm]{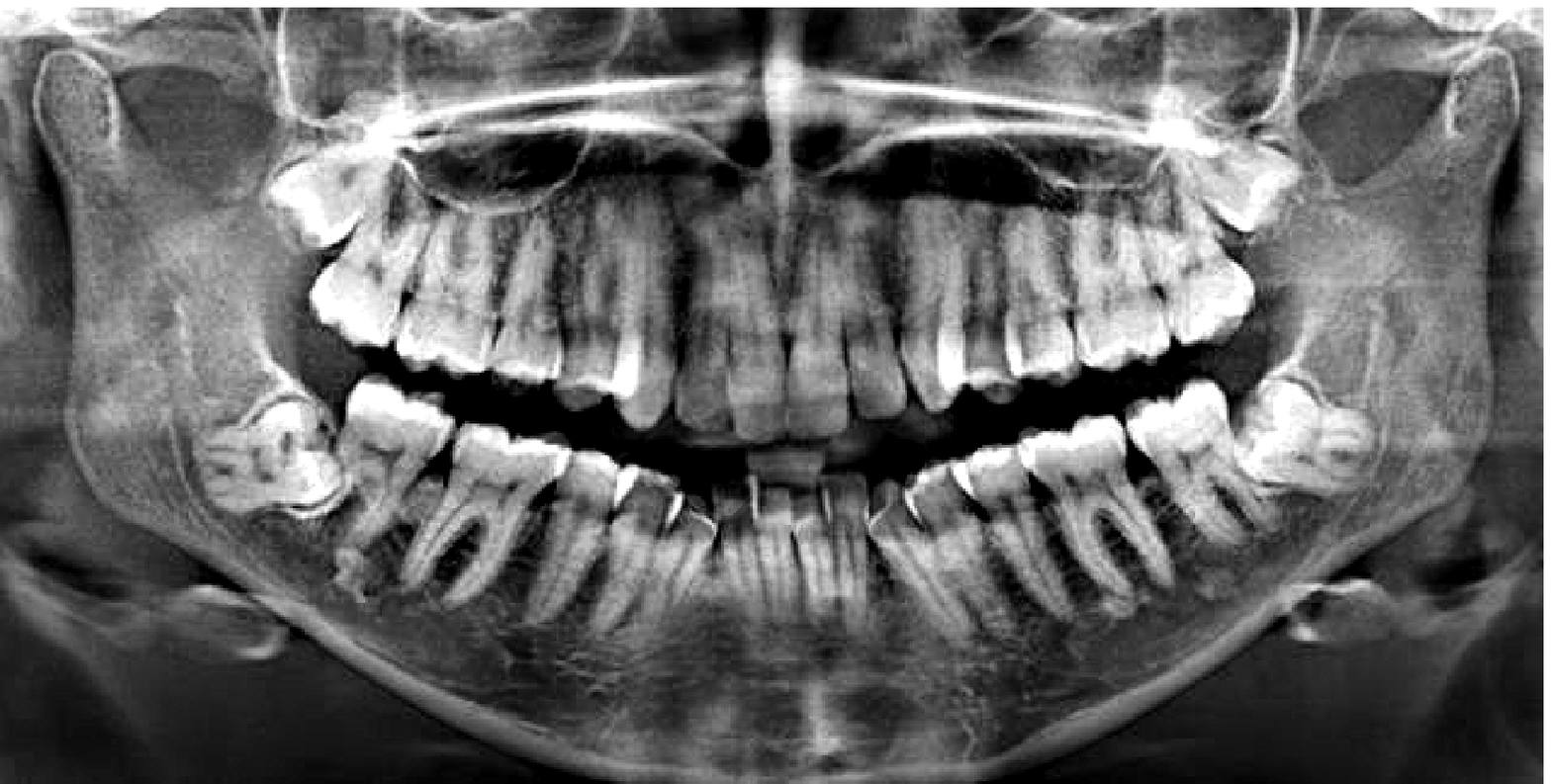}
		\caption{\label{fig:fig2}}
	\end{subfigure}
	\caption{Original dental X-ray image (a), output after applying top-hat transform (b), output after applying bottom-hat transform (c), output after applying combined equation $A_{enhance}$ (d).}
	\label{fig:figure1}
\end{figure}

To perform contrast enhancement of the image, feature extraction, and background equalization morphological top-hat and bottom-hat transform have been utilized regularly in the area of digital image processing. The result of top-hat is constructed from the difference between the input image and its opening with an SE whereas the difference between the closing by an SE and the input image represents the output of bottom-hat transform. In case of top-hat transform bright objects are fetched which are shorter than the SE. On the other hand, dark elements are retrieved with bottom-hat transform which are smaller than the SE. Therefore by combining the addition result of top-hat and subtraction result of bottom-hat with its original image, we can produce an enhanced image where the important objects will be visualized more apparently. Fig. \ref{fig:figure1} shows the individual output of top-hat, bottom-hat, and combined transform result. The equation for top-hat, bottom-hat, and enhanced image can be measured as follows where A is the input image, B is the structuring element, “$\circ$” means opening and “$\bullet$” indicates closing. \par

\begin{equation}
{A}_{top} = A - (A \circ B)
\end{equation}	

\begin{equation}
{A}_{bot} = (A \bullet B) - A
\end{equation}

\begin{equation}
{A}_{enhance} = A + {A}_{top} - {A}_{bot}
\end{equation}
\par

\subsection{Finding Optimal Structuring Element (SE) Size}

\begin{figure}[h]
	\centering
	\begin{subfigure}[b]{0.5\linewidth}
		\centering\includegraphics[width=4.2cm,height=6cm]{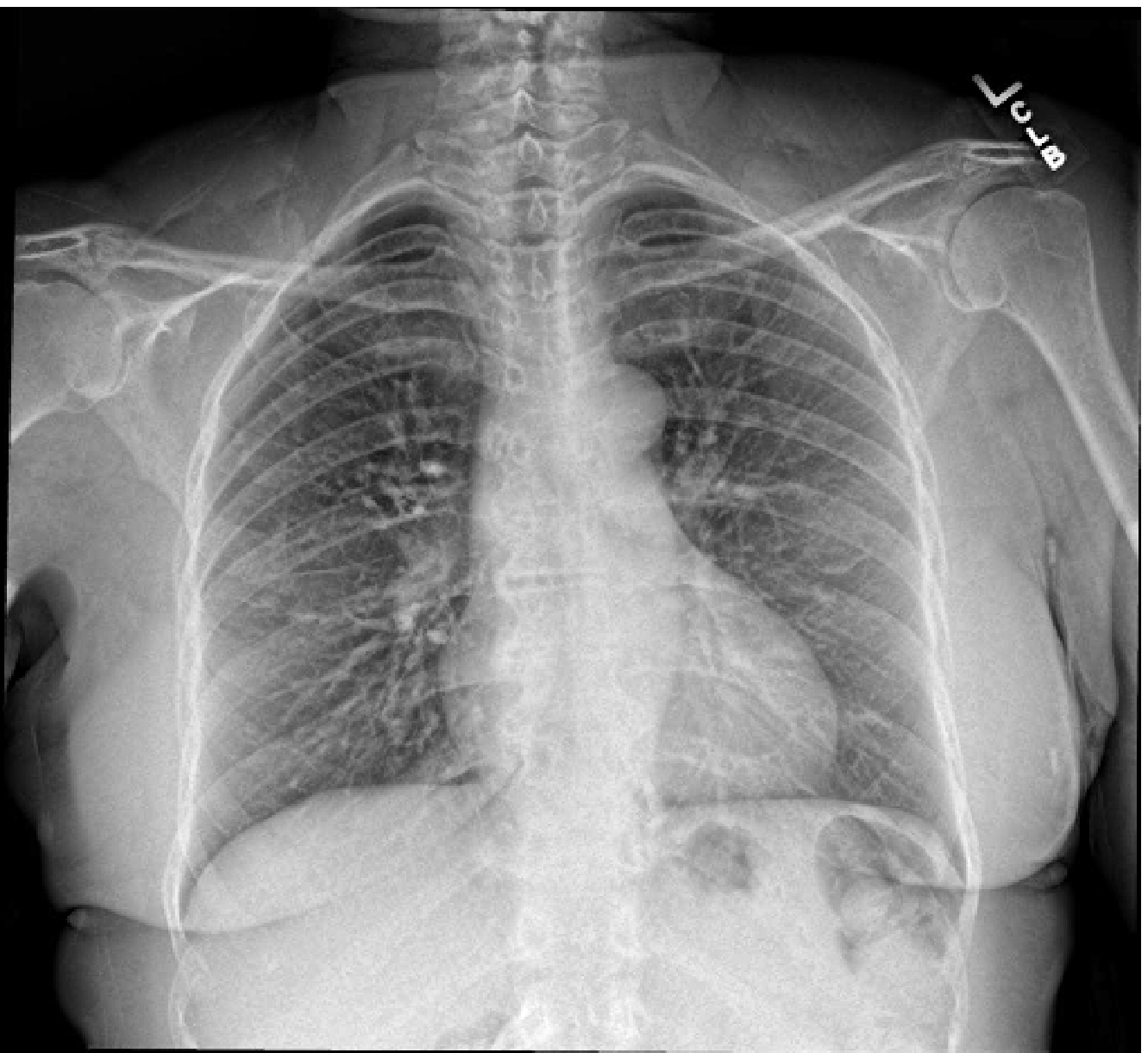}
		\caption{}
	\end{subfigure}%
	\begin{subfigure}[b]{0.5\linewidth}
		\centering\includegraphics[width=4.2cm,height=6cm]{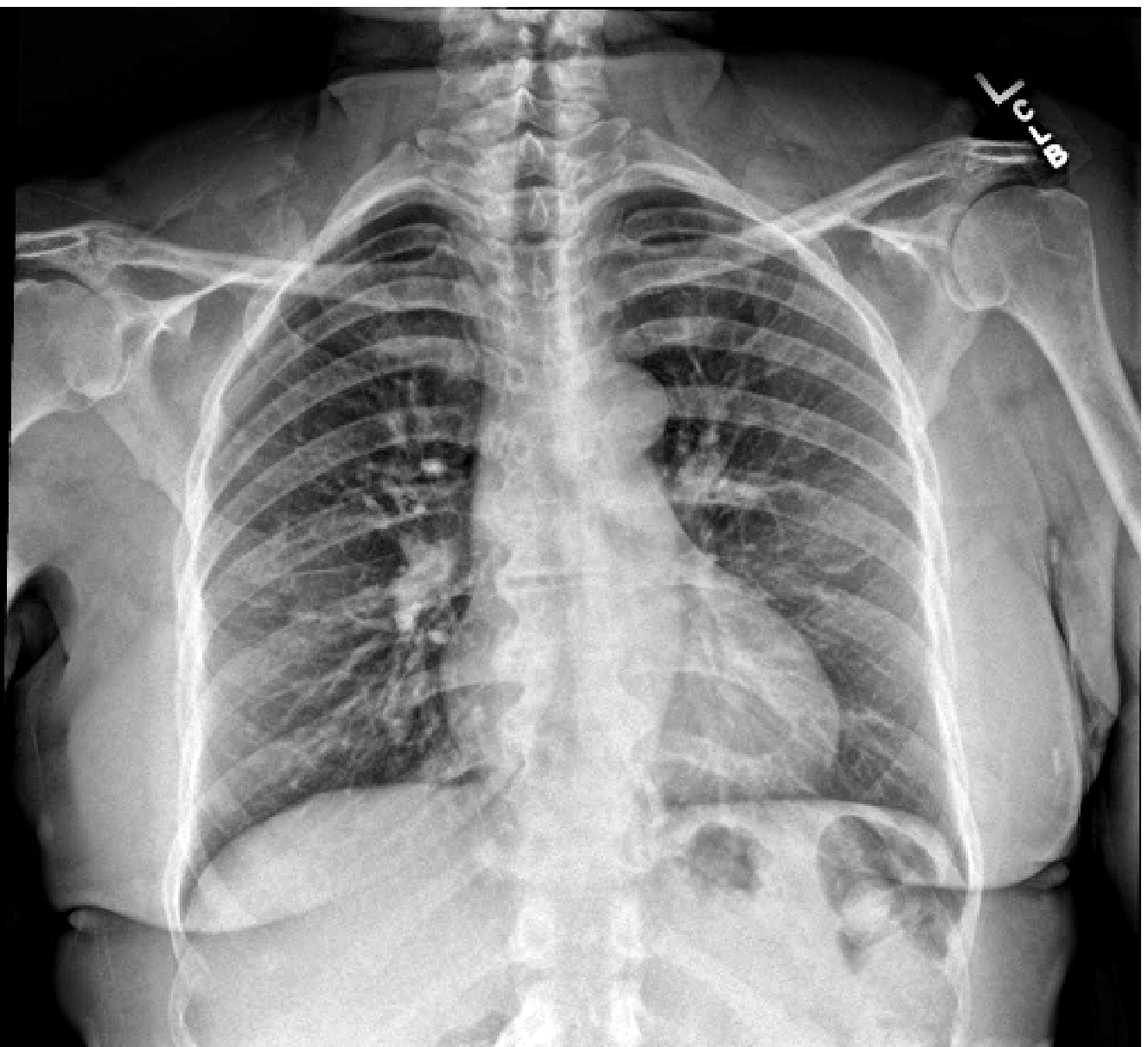}
		\caption{}
	\end{subfigure}
	\caption{Different output for taking different SE size. The size of SE is 5 for (a) and the size of SE is 15 for (b) which is more enhanced than (a).}
	\label{fig:figure2}
\end{figure}

Changing the size of the SE can significantly shift the enhancement performance. Automatic and appropriate selection of the SE size is the most challenging and important task to achieve a better result. Fig. \ref{fig:figure2} shows an example of output variation for taking different SE size of an X-ray image. Firstly, the disk type of SE shape is picked because of its rotation invariance property. Next, for automatically selecting the SE size, the process of Kushol et al. \cite{b14} is followed where Edge content (EC) based contrast matrix is measured. We start our search with a small value of SE and calculate the value of EC. Then gradually increasing the value of SE also results in an increased value in EC. The increasing behavior becomes steady after a certain amount of iterations which suggests the most contrast enhanced image possible with the given $A_{enhance}$ equation. The value of EC is basically generated from the magnitude of the gradient vector where $(x,y)$ is any pixel position of the input image A and $m \times n$ is an image block size. The following equations represent the process of measuring EC value. Here, $1 \leq x \leq m$ and $1 \leq y \leq n$. \par

\begin{equation}
\bigtriangledown A(x,y) = \begin{bmatrix}
G_x \vspace{1mm}\\ 
G_y
\end{bmatrix}
=\begin{bmatrix}
\frac{\partial }{\partial x}A(x,y)\vspace{2mm} \\ 
\frac{\partial }{\partial y}A(x,y)
\end{bmatrix}
\end{equation}

\begin{equation}
EC = \frac{1}{(m \times n)} \sum_{x} \sum_{y} \left | \bigtriangledown A(x,y) \right | 
\end{equation} 

\par

\section{EXPERIMENTAL RESULT}

The experiment is performed in MATLAB software with a system environment of 2.20 GHz processor and 8 GB RAM. At first, we have collected all publicly available benchmark X-ray image datasets to verify the efficiency of our proposed method. Different datasets focus on the different skeletal area of the body such as chest, teeth, knee etc. Furthermore, the resolution and format of the images are different from each other which help to justify the robustness of our proposed idea. \par

\subsection{ Dataset}

We have used six different datasets \cite{b15} – \cite{b19} for our experiment to test how our proposed method performs. Firstly, the Montgomery County chest X-ray set \cite{b15} is used which contains 138 frontal CXR images from Montgomery County’s Tuberculosis screening program. It includes 80 normal cases and 58 cases with manifestations of TB. Secondly, from the same work, Shenzhen chest X-ray set \cite{b15} is used, which consists of 662 frontal CXR images. Here, the number of normal cases and cases with manifestations of TB are 326 and 336 respectively. These two datasets are very well organized and comprise the patient’s age, gender, and lung abnormality information for each image. \par

Thirdly, the dataset evaluated by Wang et al. \cite{b16} is used. It contains 400 dental X-ray images in TIFF format and these were explored by certified doctors to label them. Fourthly, the JSRT dataset \cite{b17} is used which consists of 247 images of chest X-ray from 13 medical institutions in Japan. This dataset is quite old but heavily used in the field of X-ray image analysis. It has 154 CXR images with a lung nodule and rest of the 93 radiographs without a lung nodule. \par

Another great work, the ChestX-ray8 dataset \cite{b18} is tested which is the largest database we have worked with. From 32,717 unique patients, it creates 108,948 frontal view X-ray images with eight disease image labels. The eight types of diseases are – Atelectasis, Cardiomegaly, Effusion, Infiltration, Mass, Nodule, Pneumonia, and Pneumothorax. \par

Finally, a publicly available dataset provided by Demner-Fushman et al. \cite{b19} is evaluated which is published through the National Library of Medicine (NLM) image retrieval services (Open-i).  It consists of 7470 chest X-ray images both in DICOM and PNG format where most of the images are collected from the Indiana University hospital network. Table \ref{tab2} depicts the summary of the aforementioned publicly available databases with major information. \par

\begin{table} 
	\centering
	\caption{SUMMARY OF PUBLICLY AVAILABLE X-RAY IMAGE DATASET}
	\label{tab2}
	\begin{tabular}{|c|c|c|}
		\hline
		&  &    \\
		\textbf{Database Name} & \textbf{Year} & \textbf{Attributes} \\
		&  &    \\
		\hline
		 &	\multirow{5}{*}{2014} & Number of Images: 138  \\
		 \cline{3-3}
		Montgomery & & Medical Organ: Chest (Frontal) \\
		\cline{3-3}
		County chest & & Image Type: PNG \\
		\cline{3-3}
		X-ray set \cite{b15} & & Image Resolution: 4020$\times$4892 or  \\
		& & 4892$\times$4020 pixels \\
		\hline
		 & 	\multirow{4}{*}{2014} & Number of Images: 662  \\
		\cline{3-3}
		Shenzhen chest & & Medical Organ: Chest (Frontal) \\
		\cline{3-3}
		X-ray set \cite{b15}& & Image Type: PNG \\
		\cline{3-3}
		& & Image Resolution: 3K$\times$3K pixels \\
		\hline
		 & \multirow{4}{*}{2016} & Number of Images: 400  \\
		\cline{3-3}
		Dataset used by & & Medical Organ: Dental \\
		\cline{3-3}
		Wang et al. \cite{b16} & & Image Type: TIFF \\
		\cline{3-3}
		& & Image Resolution: 1935 $\times$ 2400  \\
		& &  pixels \\
		\hline
		  & 	\multirow{4}{*}{2000} & Number of Images: 247  \\
		\cline{3-3}
		JSRT Database & & Medical Organ: Chest, Lung \\
		\cline{3-3}
		\cite{b17} & & Image Resolution: 2048 $\times$ 2048 \\
		& &  pixels \\
		\hline
		\multirow{4}{*}{ChestX-ray8 \cite{b18}} & 	\multirow{4}{*}{2017} & Number of Images: 108948  \\
		\cline{3-3}
		& & Medical Organ: Chest (Frontal) \\
		\cline{3-3}
		& & Image Resolution: 2000 $\times$ 3000  \\
		& &  pixels \\
		\hline
		\multirow{3}{*}{NLM (Open-i)}  & 	\multirow{3}{*}{2015} & Number of Images: 7470  \\
		\cline{3-3}
		\multirow{3}{*}{Dataset \cite{b19}}& & Medical Organ: Chest  \\
		\cline{3-3}
		& & Image Type: PNG and DICOM \\
		\hline
	\end{tabular}
\end{table}

\subsection{Comparison with another method}

\begin{figure}[h]
	\centering	
	\begin{subfigure}[b]{0.5\linewidth}
		\centering\includegraphics[width=6cm,height=7cm]{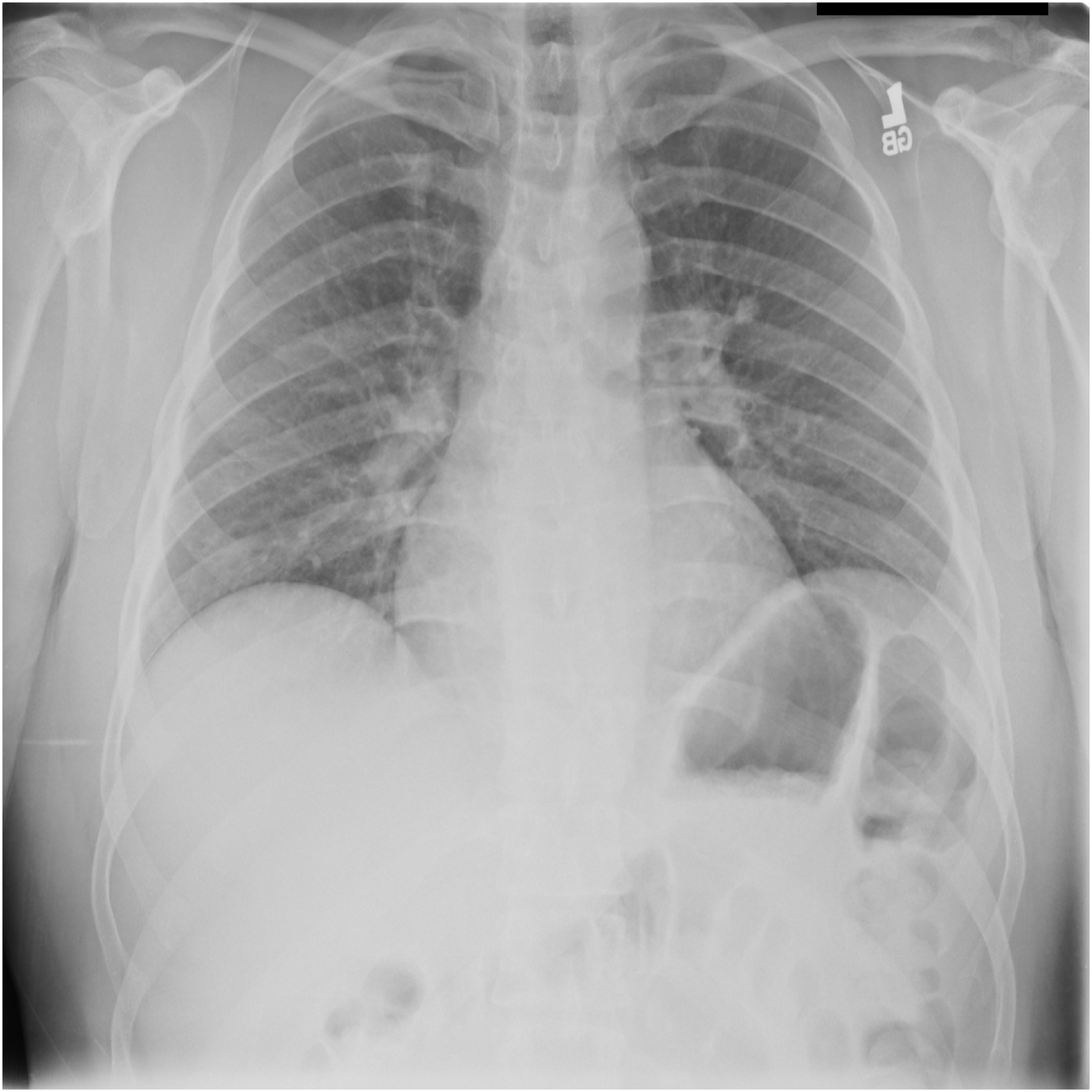}
		\caption{}
	\end{subfigure} \\
	\begin{subfigure}[b]{0.5\linewidth}
		\centering\includegraphics[width=6cm,height=7cm]{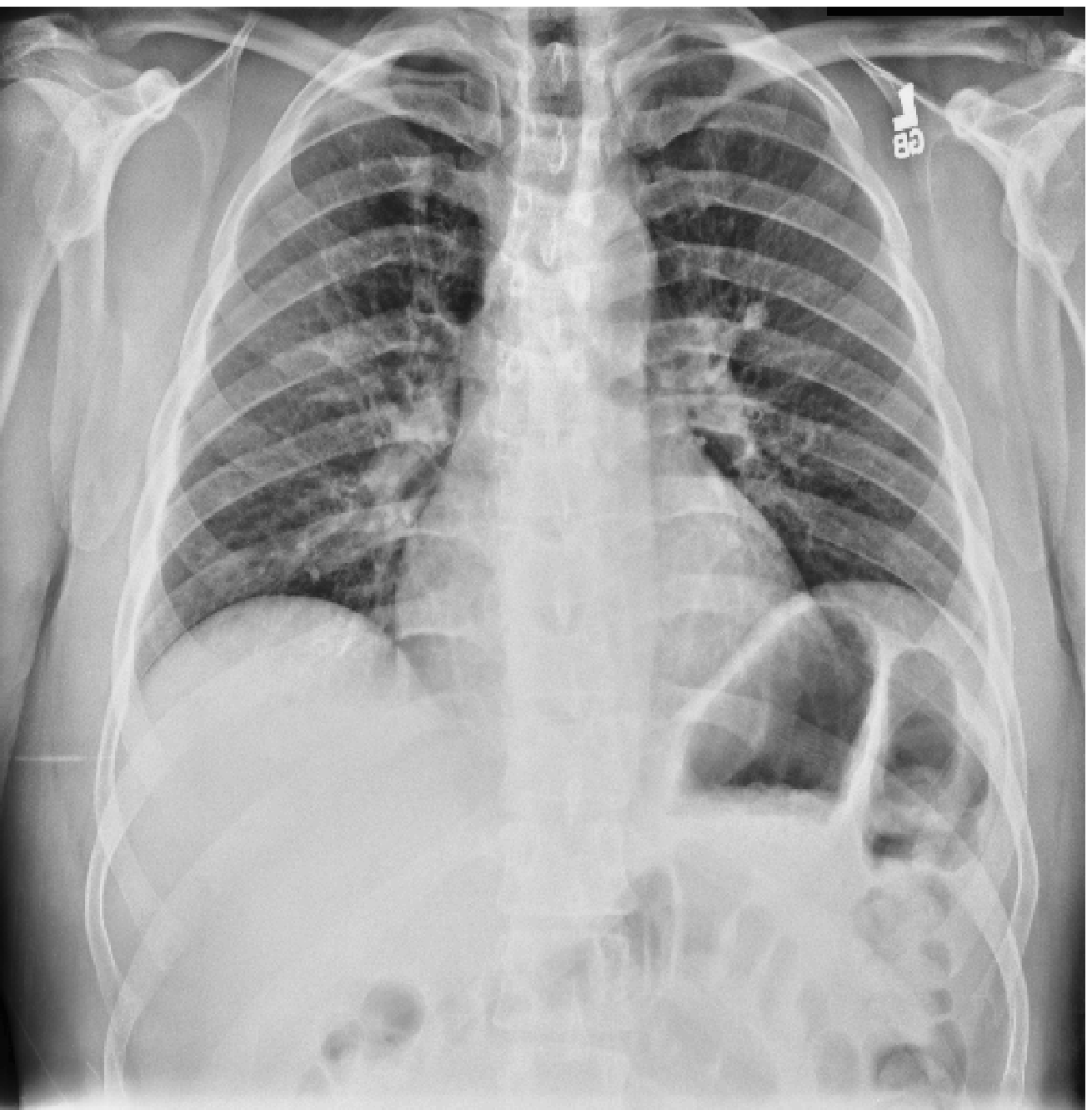}
		\caption{}
	\end{subfigure} \\ 
	\begin{subfigure}[b]{0.5\linewidth}
		\centering\includegraphics[width=6cm,height=7cm]{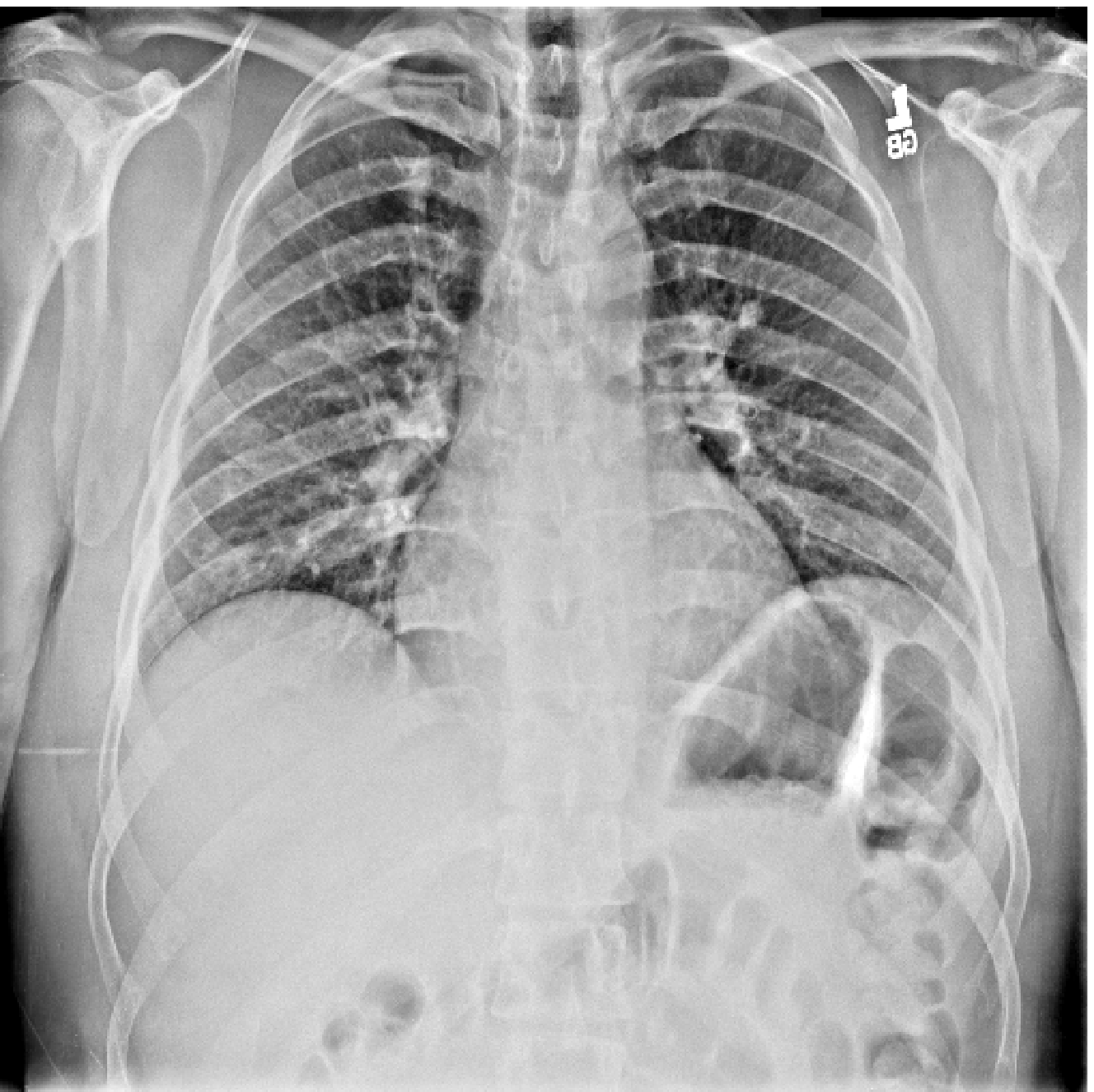}
		\caption{}
	\end{subfigure}
	\caption{CXR image taken from ChestX-ray8 \cite{b18}, (a) original image, (b) CLAHE output, (c) proposed method output.}
	\label{fig:figure3}
\end{figure}

\begin{figure}[h]
	
	\centering
	\begin{subfigure}[b]{0.5\linewidth}
		\centering\includegraphics[width=6cm,height=7cm]{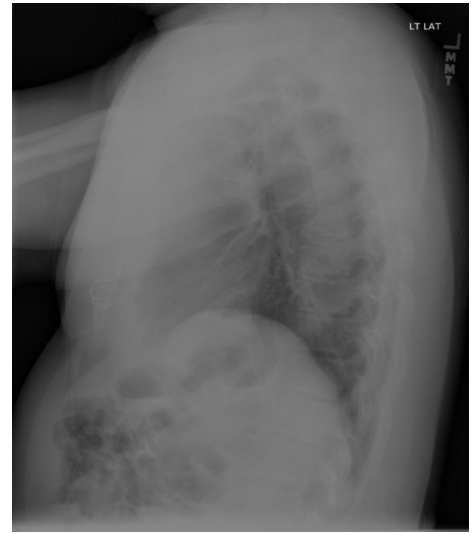}
		\caption{}
	\end{subfigure} \\
	\begin{subfigure}[b]{0.5\linewidth}
		\centering\includegraphics[width=6cm,height=7cm]{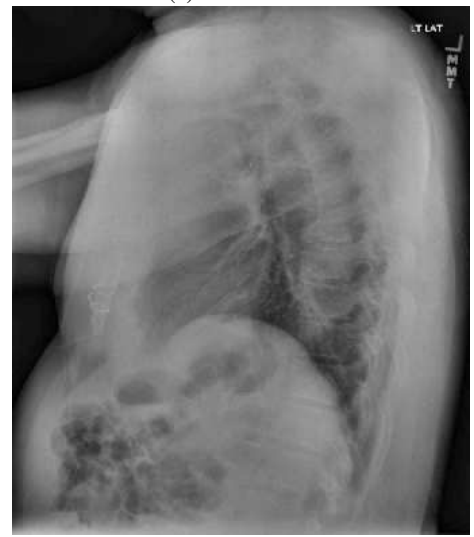}
		\caption{}
	\end{subfigure} \\
	\centering
	\begin{subfigure}[b]{0.5\linewidth}
		\centering\includegraphics[width=6cm,height=7cm]{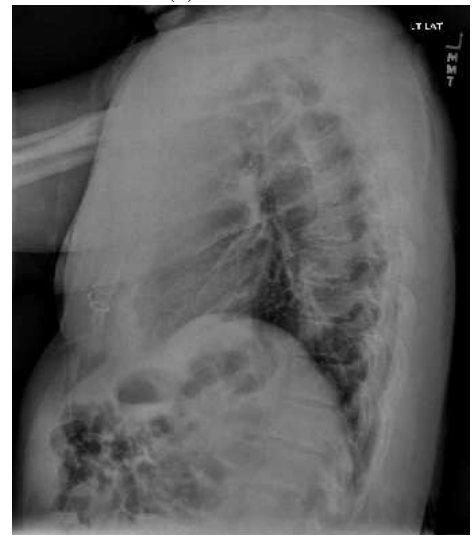}
		\caption{}
	\end{subfigure}
	\caption{X-ray image taken from NLM (Open-i) Dataset \cite{b19} (a) original image, (b) CLAHE output, (c) proposed method output.}
	\label{fig:figure4}
\end{figure}

We have compared our proposed method with one prominent image enhancement method CLAHE which is frequently used to enhance different types of medical images. The size of tiles by row and column was $8\times8$ whereas the number of bins for the histogram was 128 to implement the CLAHE method on MATLAB. Fig. \ref{fig:figure3} and Fig. \ref{fig:figure4} show the comparison of the output result of our method with CLAHE where the sample images are taken from ChestX-ray8 \cite{b18} and NLM (Open-i) dataset \cite{b19} respectively. In both of the images, we can observe that our proposed method generates more visually improved and clear output result which will eventually help the physicians to correctly identify the proper disease. \par

\section{Conclusion}

To speed up the diagnosis process in the medical health community, image enhancement plays an important role. Time consumption can be reduced significantly by assisting doctors with a clear image view. However, accuracy in correct disease detection is the most vital factor which can be increased through contrast enhanced image. The method proposed in this paper can automatically enhance the contrast of different X-ray images even if the resolution of the image is increased. Our future work can be segmentation of a specific region of interest for the identification of specific disease. For example, Heart, lung, clavicles etc. can be segmented from a clear chest X-ray image.

\end{document}